\documentclass[11pt,a4paper]{article}
\usepackage[T2A,T1]{fontenc}
\usepackage[utf8]{inputenc}
\usepackage[hyperref]{acl2019}
\usepackage{times}
\usepackage{latexsym}
\usepackage{booktabs}
\usepackage{multirow}
\usepackage{graphicx}
\usepackage{isomath}
\usepackage{amsmath,amssymb}
\usepackage{url}
\usepackage{relsize}
\usepackage{todonotes}

\aclfinalcopy 
\def\aclpaperid{40} 

\title{One-to-X analogical reasoning on word embeddings: \\ a case for diachronic armed conflict prediction from news texts}

\author{Andrey Kutuzov \\
  University of Oslo \\
  Oslo, Norway \\
  \texttt{andreku@ifi.uio.no} \\\And
  Erik Velldal \\
  University of Oslo \\
  Oslo, Norway \\
  \texttt{erikve@ifi.uio.no} \\\And
  Lilja Øvrelid \\
  University of Oslo \\
  Oslo, Norway \\
  \texttt{liljao@ifi.uio.no}}

\date{}

\begin{document}
\maketitle
\begin{abstract}
    We extend the well-known word analogy task to a one-to-X formulation, including one-to-none cases, when no correct answer exists. The task is cast as a relation discovery problem and applied to historical armed conflicts datasets, attempting to predict new relations of type `location:armed-group' based on data about past events. As the source of semantic information, we use diachronic word embedding models trained on English news texts. 
    A simple technique to improve diachronic performance in such task is demonstrated, using a threshold based on a function of cosine distance to decrease the number of false positives; this approach is shown to be beneficial on two different corpora. Finally, we publish a ready-to-use test set for one-to-X analogy evaluation on historical armed conflicts data.
\end{abstract}

Performance on the task of analogical inference (or `word analogies') is one of the most widespread means to evaluate distributional word representation models, with `\texttt{KING} is to \texttt{QUEEN} as \texttt{MAN} is to \texttt{?} (\texttt{WOMAN})' being a famous example. It also has deep connections to the relational similarity task \cite{semeval:2012}.
Most often, analogical inference is formulated as a strict proportion, and the model has to provide exactly one best answer for each question (assuming that it is impossible that, e.g., \texttt{WOMAN} and \texttt{GIRL} are equally correct answers for the question above).

We reformulate the analogical inference task and extend it to include multiple-ended or \textit{one-to-X} relations: one-to-one, one-to-many and one-to-none cases when an entity is not included in this particular relation type, so there is no correct answer for it. This way, the model has to provide as many correct answers as possible, while providing as few incorrect answers as possible. More formally, the task is as follows: for a given vocabulary $V$, a relation of a type $z$, and an entity $x \in V$, identify any pairs $x;i \in V$ such that $z$ holds between $x$ and $i$. Note that this task has been tackled in NLP using a number of methods, and not necessarily using analogical reasoning; however, in this work we employ a supervised approach implying learning from `example' or `prototypical' pairs (similar to analogies). Our method also does not require providing $i$ candidates: they are inferred automatically from an embedding model.

Proper analogy test sets are difficult to compile, especially when the complex structure described above is desired. Thus, we limit ourselves to one particular type of semantic relations, on which objective data can be gathered from extra-linguistic sources: those between a geographical \textit{location} (country) and an \textit{insurgent group} involved in an armed conflict against the government of the country in a given time period. We use the historical armed conflicts data provided publicly by the UCDP project \cite{gleditsch2002armed}. These datasets contain the needed relations: several armed groups can operate in one location, one group can operate in several locations, and obviously some locations lack any insurgents to speak of. At the same time, news corpora contain a lot of information about armed conflicts, while being  comparatively easy to obtain and train distributional word embedding models on.

Since the UCDP data provides exact dates for all the conflicts, we cast our \textit{one-to-X} analogical reasoning task in a diachronic setup. We attempt to find out whether a distributional vector space retains enough structure to trace the relation after the model was additionally trained with a comparable amount of new in-domain texts created in the subsequent time period. 

The \textbf{contributions} of this work are: \textbf{(1)} We reformulate the well-known word analogy task such that multiple correct answers or no correct answer at all become possible (\textit{one-to-X} relations). \textbf{(2)} We process historical armed conflicts data and present it as a ready-to-use evaluation set.
\textbf{(3)} Relying on and partially reproducing the workflow from prior publications, we
investigate whether word embedding models are able to solve \textit{one-to-X} analogies diachronically.
\textbf{(4)} Finally, we show that our learned cosine threshold approach can significantly improve the temporal \textit{one-to-X} analogies performance by filtering out false positives.

\section{Related work} \label{sec:related}
The issue of linguistic regularity manifested in relational similarity has been studied for a long time. Due to the long-standing criticism of strictly binary relation structure,  \textit{SemEval-2012} offered the task to detect the degree of relational similarity \cite{semeval:2012}. This means that multiple correct answers exist, but they should be ranked differently. Somewhat similar improvements to the well-known word analogies dataset from \cite{Mikolov_representation:2013} were presented in the BATS analogy test set \cite{GladkovaDrozd2016}, also featuring multiple correct answers.\footnote{See also the detailed criticism of analogical inference with word embeddings in general in \cite{rogers:analogy:2017}.} Our \textit{One-to-X} analogy setup extends this by introducing the possibility of the correct answer being 'None'. In the cases when correct answers exist, they are equally ranked, but their number can be different.

Using distributional word representations to trace diachronic semantic shifts (including those reflecting social and cultural events) has received substantial attention in the recent years. Our work shares some of the workflow with \newcite{kutuzov:relations}. They used a supervised approach to analogical reasoning, applying `semantic directions' learned on the previous year's armed conflicts data to the subsequent year. We extend their research by significantly reformulating the analogy task, making it more realistic, and finding ways to cope with false positives (insurgent armed groups predicted for locations where no armed conflicts are registered this year). In comparison to their work, we also use newer and larger corpora of news texts and the most recent version of the UCDP dataset.
For brevity, we do not describe the emerging field of diachronic word embeddings in details, referring the interested readers to the recent surveys of \newcite{kutuzov_survey} and \newcite{tang_2018}.

\section{Learning the armed conflict projection} \label{sec: projection}
We rely on the idea that knowing the gold \textit{Location: Insurgent} pairs from a time period $n$ can help us to retrieve the correct pairs bearing the same relation from the next time period $n+1$, using word embedding models trained incrementally\footnote{The model $M_{n+1}$ is initialized with the weights from the model $M_{n}$; if there are new words in the $n+1$ data which exceed the frequency threshold, then at the start of $M_{n+1}$ training they are added to it and assigned random vectors.} on these time periods. The models are trained using the CBOW algorithm \cite{Mikolov_representation:2013}, and the time periods are yearly subsections of English news corpora (see \S~\ref{sec: data}). A yearly model is saved after the training for a particular year is finished, for later usage.

We deal with pairs of consequent years (`2010--2011', `2011--2012', etc.). Our aim is to predict armed conflicts (or their absence) for a fixed set of locations in the year $n+1$. Having the gold armed conflict data for all years, we can train a predictor on the 1\textsuperscript{st} year, and then evaluate it on the 2\textsuperscript{nd} one (simulating a real-world scenario where new textual data arrive regularly, but gold annotation is available only for older data).
We take the gold \textit{Location: Insurgent} pairs from the year $n$ (as a rule, there are several dozens of them) and their vector representations from the corresponding embedding model $M_{n}$. Then, these vector pairs are used to train a linear projection $\vectorsym{T} \in \mathbb{R}^{p \times d}$, where $p$ is the number of pairs, and $d$ is the vector size of the embedding model used. 

Linguistically, $\vectorsym{T}$ can be seen as defining a `prototypical armed conflict relation'; geometrically, it can be thought of as the average `direction' from locations to their active insurgent groups in the $M_{n}$ vector space \footnote{A similar approach has been used for naive translation of words from the language L1 to L2 by using monolingual word embeddings for both and a seed bilingual dictionary (set of one-to-one pairs) \cite{mikolov2013translation}.}. 
The problem of finding the optimal $\vectorsym{T}$ boils down to a linear regression which minimizes the error in transforming one set of vectors into another, and we do it by solving $d$ deterministic normal equations (since the number of data points is small, the operation is fast). 

After $\vectorsym{T}$ is at hand, one can find the `armed conflict projection' vector $\hat{i}$ for any  location vector $\vectorsym{v}$  in $M_{n+1}$ by transforming it with the learned matrix: $\hat{i} = \vectorsym{v} \cdot \vectorsym{T}$. In the simplest case, the word with the highest cosine similarity to $\hat{i}$ in $M_{n+1}$ is assumed to be a candidate for an insurgent armed group active in this location in the time period $n+1$; however, a more involved approach is needed to handle cases when the number of insurgents (correct answers) can be different from 1 (including 0), described below. 

For this workflow to yield meaningful results, it is essential for the paired models to be `aligned'.
This is why we train the models incrementally, thus ensuring that they share common structural properties. Another possible way to cope with this is by using the orthogonal Procrustes alignment \cite{hamilton2016cultural}.

\section{Datasets} \label{sec: data}
\paragraph{Corpora for embeddings} We train embeddings 
on two corpora: 
\textbf{(1)} The \textit{Gigaword} news corpus \cite{Gigaword:11}, spanning 1995--2010 and containing about 300M words per year, with about 4.8 billion total. This corpus was used in \cite{kutuzov:relations} and we include it for comparison purposes.
\textbf{(2)} The \textit{News on Web (NOW)} corpus,\footnote{\url{https://corpus.byu.edu/now/}}  spanning 2010--2019. As the UCDP dataset covers conflicts only up to 2017, we use the texts up to that year, yielding on average 730M words per year, with about 5.9 billion total. The time-annotated texts are crawled from online magazines and newspapers in 20 English-speaking countries.

Before training the embedding models, the corpora were lemmatized and PoS-tagged using \textit{UDPipe 2.3 English-LinES} tagger \cite{udpipe:2017} (during the evaluation, PoS tags were stripped and words lower-cased). Chains of consecutive proper names agreeing in number (\texttt{`South\_PROPN Sudan\_PROPN'}) were merged together with a special character (\texttt{`South::Sudan\_PROPN'}). This was important to handle multi-word location and insurgent names. Functional words were removed.

\paragraph{Conflict relation data} 
The armed conflict data comes from the UCDP/PRIO Armed Conflict Dataset\footnote{\url{https://www.ucdp.uu.se/}} (ver. 18.1)  \cite{pettersson2018organized}. It is manually annotated with historical information on armed conflicts across the world, 
starting from 1946, where at least one party is the government of a state, 
and frequently used in statistical conflict research. 

The dataset contains various metadata, but we kept only the years, the names of the locations, and the names of the armed groups (e.g., \texttt{`2016: Afghanistan: ["Taliban", "Islamic State"]'}). Entities occurring less than 25 times in the corresponding yearly corpora were filtered out, since it is difficult for distributional models to learn meaningful embeddings for such rare words.

We create one such conflict relation dataset for each news corpus; one corresponding to the time span of NOW and another for Gigaword. Table~\ref{tab:datasets} shows various statistics across these UCDP subsets, including the important `new pairs share' parameter, showing what part of the conflict pairs in the years $n+1$ was not seen in the years $n$ (how much new data to guess).

The \textit{NoW} dataset features 102 unique \textit{Location: Insurgent} pairs, with 42 unique locations and 78 unique armed groups. On average, each year 56\% of these 42 locations were involved in armed conflicts, based on the UCDP data. The remaining (different each year) serve as `negative examples' to test the ability of our approach to detect cases when no predictions have to be made. For the areas involved in conflicts, the average number of active insurgents per location is about 1.5, with the maximum number being 5\footnote{Congo (2017) features 5 active armed groups: \texttt{`Kamuina Nsapu'}, \texttt{`M23'}, \texttt{`CMC'}, \texttt{`MNR'}, \texttt{`BDK'}.}.

\begin{table}
\smaller
\center
\begin{tabular}{lcc}
\toprule
&\textbf{Gigaword}&\textbf{NOW} \\
\midrule
Time span &1995--2010 &2010--2017 \\
Locations & 52 & 42 \\
Insurgents & 127 & 78 \\
Conflict pairs & 136 & 102 \\
New pairs share & 0.37 & 0.39 \\
Conflict locations share & 0.46 & 0.56  \\
Insurgents per location & 1.65 & 1.50 \\
\bottomrule
\end{tabular}
\caption{Comparative statistics of UCDP data subsets}
\label{tab:datasets}
\end{table}

\paragraph{A replication experiment}  
In Table~\ref{tab:eval_old} we replicate the experiments from \cite{kutuzov:relations} on both sets. It follows their evaluation scheme, where only the presence of the correct armed group name in the 
$k$ 
nearest neighbours of the $\hat{i}$ mattered, and only conflict areas were present in the yearly test sets. Essentially, it measures the recall 
$@k$, 
without penalizing the models for yielding incorrect answers along with the correct ones, and never asking questions having no correct answer at all (e.g., peaceful locations). The performance is very similar on both sets, ensuring that  the \textit{NOW} set conveys the same signal as the \textit{Gigaword} set; however, in the next section we make the task more realistic by extending the evaluation schema to the \textit{one-to-X} scenario described above.

\begin{table}
\smaller
\center
\begin{tabular}{lccc}
\toprule
\textbf{Dataset} &\textbf{ @1 }  &\textbf{@5}    & \textbf{@10}\\
\midrule
Gigaword & 0.356 &0.555 & 0.610\\
NOW  & 0.442 & 0.557 & 0.578\\
\bottomrule
\end{tabular}
\caption{Average recall of diachronic analogy inference}
\label{tab:eval_old}
\end{table}

\section{Evaluation setup} \label{sec: eval}
In our workflow, each yearly test set contains all locations, but whether a particular location is associated with any armed groups, can vary from year to year. Conceptually, the task of the model is to predict correct sets of active armed groups for conflict locations and to predict the empty set for peaceful locations. For a test year, an `armed conflict projection' $\hat{i}$ is produced for each location using the learned transformation $\vectorsym{T_n}$. 
The $k$ 
nearest neighbors of $\hat{i}$ in $M_{n+1}$ become armed group candidates 
($k$ is a hyperparameter). 
We calculate the number of true positives (correctly predicted armed groups), false positives (incorrectly predicted armed groups), and false negatives (armed groups present in the gold data, but not predicted by the system). These counts are accumulated and for each year standard precision, recall and F1 score are calculated. These metrics are then averaged across all years in the test set. Using false negatives ensures that we penalize the systems for yielding predictions for peaceful locations.

\subsection{Cosine threshold}
It is clear that such a system (dubbed `baseline') will always yield 
$k$  
incorrect candidates for peaceful areas. 
Inspired partially by the ideas from \newcite{orlikowski:2018}, we implemented a simple approach based on the assumption that the correct armed groups vectors will tend to be closer to the $\hat{i}$ point than other nearest neighbours. Thus, the system should pick only the candidates located within a hypersphere of a pre-defined radius $r$ centered around $\hat{i}$. $r_n$ can be different for different years, and we infer it from the $p$ training conflict pairs from the previous year by calculating the average cosine distance between the `armed conflict projections' $\hat{i}$ and armed groups:
\begin{equation}
r = \frac{1}{p} \sum_{p=0}^p \cos\left(\hat{i}_p, g_p\right) +  \sigma
\end{equation} 
where $g_p$ is the armed group in the p\textsuperscript{th} pair, and $\sigma$ is one standard deviation of the cosine distances in $p$.
The hypersphere serves as a cosine threshold.

\begin{figure}
    \centering
    \includegraphics[scale=0.5,keepaspectratio]{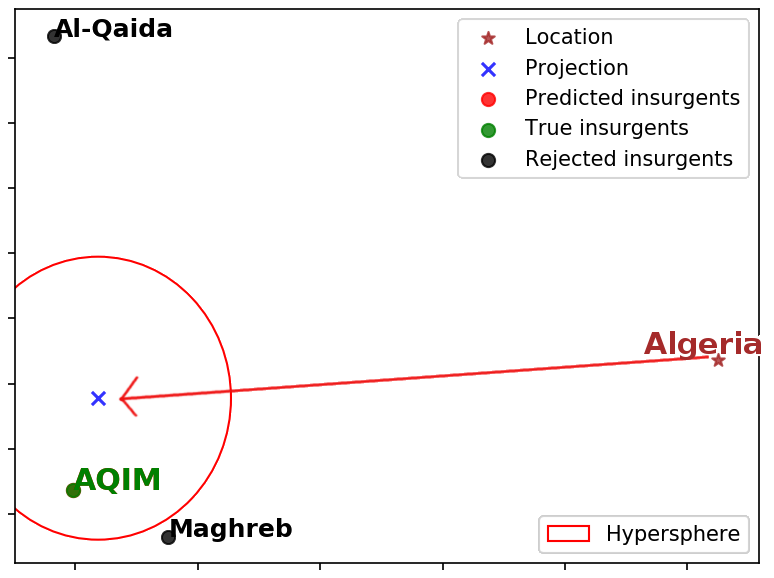}
    \caption{Prediction of armed groups in Algeria, 2014 (2-dimensional PCA projection).}
    \label{fig:yemen}
\end{figure}

This allows us to keep only the candidates which are not farther from $\hat{i}$ than the armed groups in the previous year tended to be. Figure \ref{fig:yemen} shows a PCA projection of predicting armed groups for Algeria in 2014. With 
$k=3$, 
the system initially yielded 3 candidates (\texttt{`AQIM'}, \texttt{`Al-Qaida'} and \texttt{`Maghreb'}), with only the first being correct. The red circle is a part of the hypersphere inferred from the 2013 training data. It filters out the wrong candidates (in black), since the cosine distance from the conflict projection (in blue) to their embeddings is higher than the inferred threshold.

\section{Experiments} \label{sec: experiments}
For the experiments, we chose 
$k=2$, 
to be closer to the average number of armed groups per location in  our sets. 
Table~\ref{tab:eval_new_diachronic} shows the \textit{diachronic} performance of our system in the setup when the matrix $\vectorsym{T}_n$ and the threshold $r_n$ are applied to the year $n+1$. 

\begin{table}
\center
\begin{tabular}{llccc}
\toprule
& \textbf{Algorithm} &\textbf{Precision} & \textbf{Recall} & \textbf{F1} \\
\midrule
\multirow{2}{*}{\rotatebox{90}{Giga}}
&Baseline & 0.19 & 0.51 & 0.28 \\
&Threshold & 0.46 & 0.41 & \textbf{0.41}  \\
\midrule
\multirow{2}{*}{\rotatebox{90}{NOW}}
&Baseline & 0.26 & 0.53 & 0.34 \\
&Threshold & 0.42 & 0.41 & \textbf{0.41}  \\
\bottomrule
\end{tabular}
\caption{Average diachronic performance}
\label{tab:eval_new_diachronic}
\end{table}

For both \textit{Gigaword} and  \textit{NOW} datasets (and the corresponding embeddings), using the cosine-based threshold decreases recall and increases precision (differences are statistically significant with \textit{t-test}, $p  < 0.05$). At the same time, the integral metrics of F1 consistently improves ($p < 0.01$). Thus, the thresholding reduces prediction noise in the \textit{one-to-X} analogy task without sacrificing too many correct answers. In our particular case, this helps to more precisely detect events of armed conflicts termination (where no insurgents should be predicted for a location), not only their start. 

As a sanity check, we also evaluated it \textit{synchronically}, that is when $\vectorsym{T}_n$ and $r_n$ are tested on the locations from the same year (including peaceful ones). In this easier setup, we observed exactly the same trends (Table \ref{tab:eval_new_synchronic}). 

\begin{table}
\center
\begin{tabular}{llccc}
\toprule
&\textbf{Algorithm} &\textbf{Precision} & \textbf{Recall} & \textbf{F1} \\
\midrule
\multirow{2}{*}{\rotatebox{90}{Giga}}
&Baseline & 0.28 & 0.74 & 0.41 \\
&Threshold & 0.60 & 0.69 & \textbf{0.63}  \\
\midrule
\multirow{2}{*}{\rotatebox{90}{NOW}}
&Baseline & 0.39  & 0.88 & 0.53 \\
&Threshold & 0.50  & 0.77 & \textbf{0.60}  \\
\bottomrule
\end{tabular}
\caption{Average synchronic performance}
\label{tab:eval_new_synchronic}
\end{table}

\section{Conclusion} \label{sec:conclusion}
We presented a new \textit{one-to-X} word analogy task formulation, applying it to the problem of temporal armed conflicts detection based on word embedding models trained on English news texts. A historical armed conflicts test set was prepared for evaluation of diachronic word embedding models. We also showed that a simple thresholding technique based on a function of cosine distance allows us to significantly improve the relation detection performance, especially for reducing the number of false positives. This approach outperformed the baseline both with the corpora used in the prior work (\textit{Gigaword}) and with the \textit{NOW} corpus which to the best of our knowledge was not used for diachronic semantic shifts research before. 

Our future plans include using negative sampling when calculating optimal projections, along with testing recent diachronic modeling algorithms representing time as a continuous variable \cite{rosenfeld:2018}. Another interesting issue is how to avoid catastrophic forgetting when training embeddings incrementally (semantic relation structures tend to completely change after significant updates).

Our code, test sets and best-performing embeddings are available at \url{https://github.com/ltgoslo/diachronic_armed_conflicts}.

\bibliography{acl2019}

\begin{thebibliography}{15}
\expandafter\ifx\csname natexlab\endcsname\relax\def\natexlab#1{#1}\fi

\bibitem[{Gladkova et~al.(2016)Gladkova, Drozd, and
  Matsuoka}]{GladkovaDrozd2016}
Anna Gladkova, Aleksandr Drozd, and Satoshi Matsuoka. 2016.
\newblock \href {https://doi.org/10.18653/v1/N16-2002} {Analogy-based detection
  of morphological and semantic relations with word embeddings: What works and
  what doesn't}.
\newblock In \emph{Proceedings of the NAACL-HLT SRW}, pages 47--54, San Diego,
  California, June 12-17, 2016. ACL.

\bibitem[{Gleditsch et~al.(2002)Gleditsch, Wallensteen, Eriksson, Sollenberg,
  and Strand}]{gleditsch2002armed}
Nils~Petter Gleditsch, Peter Wallensteen, Mikael Eriksson, Margareta
  Sollenberg, and H{\aa}vard Strand. 2002.
\newblock Armed conflict 1946-2001: A new dataset.
\newblock \emph{Journal of peace research}, 39(5):615--637.

\bibitem[{Hamilton et~al.(2016)Hamilton, Leskovec, and
  Jurafsky}]{hamilton2016cultural}
William Hamilton, Jure Leskovec, and Dan Jurafsky. 2016.
\newblock \href {https://doi.org/10.18653/v1/D16-1229} {Cultural shift or
  linguistic drift? {C}omparing two computational measures of semantic change}.
\newblock In \emph{Proceedings of the Conference on Empirical Methods in
  Natural Language Processing}, pages 2116--2121, Austin, Texas.

\bibitem[{Jurgens et~al.(2012)Jurgens, Mohammad, Turney, and
  Holyoak}]{semeval:2012}
David Jurgens, Saif Mohammad, Peter Turney, and Keith Holyoak. 2012.
\newblock \href {http://aclweb.org/anthology/S12-1047} {Semeval-2012 task 2:
  Measuring degrees of relational similarity}.
\newblock In \emph{*SEM 2012: The First Joint Conference on Lexical and
  Computational Semantics -- Volume 1: Proceedings of the main conference and
  the shared task, and Volume 2: Proceedings of the Sixth International
  Workshop on Semantic Evaluation (SemEval 2012)}, pages 356--364. Association
  for Computational Linguistics.

\bibitem[{Kutuzov et~al.(2018)Kutuzov, {\O}vrelid, Szymanski, and
  Velldal}]{kutuzov_survey}
Andrey Kutuzov, Lilja {\O}vrelid, Terrence Szymanski, and Erik Velldal. 2018.
\newblock \href {http://aclweb.org/anthology/C18-1117} {Diachronic word
  embeddings and semantic shifts: a survey}.
\newblock In \emph{Proceedings of the 27th International Conference on
  Computational Linguistics}, pages 1384--1397. Association for Computational
  Linguistics.

\bibitem[{Kutuzov et~al.(2017)Kutuzov, Velldal, and
  {\O}vrelid}]{kutuzov:relations}
Andrey Kutuzov, Erik Velldal, and Lilja {\O}vrelid. 2017.
\newblock \href {http://aclweb.org/anthology/D17-1194} {Temporal dynamics of
  semantic relations in word embeddings: an application to predicting armed
  conflict participants}.
\newblock In \emph{Proceedings of the Conference on Empirical Methods in
  Natural Language Processing}, pages 1824--1829, Copenhagen, Denmark.

\bibitem[{Mikolov et~al.(2013{\natexlab{a}})Mikolov, Le, and
  Sutskever}]{mikolov2013translation}
Tomas Mikolov, Quoc Le, and Ilya Sutskever. 2013{\natexlab{a}}.
\newblock Exploiting similarities among languages for machine translation.
\newblock ArXiv preprint arXiv:1309.4168.

\bibitem[{Mikolov et~al.(2013{\natexlab{b}})Mikolov, Sutskever, Chen, Corrado,
  and Dean}]{Mikolov_representation:2013}
Tomas Mikolov, Ilya Sutskever, Kai Chen, Greg~S Corrado, and Jeff Dean.
  2013{\natexlab{b}}.
\newblock Distributed representations of words and phrases and their
  compositionality.
\newblock \emph{Advances in Neural Information Processing Systems},
  26:3111--3119.

\bibitem[{Orlikowski et~al.(2018)Orlikowski, Hartung, and
  Cimiano}]{orlikowski:2018}
Matthias Orlikowski, Matthias Hartung, and Philipp Cimiano. 2018.
\newblock \href {http://aclweb.org/anthology/W18-4501} {Learning diachronic
  analogies to analyze concept change}.
\newblock In \emph{Proceedings of the Second Joint SIGHUM Workshop on
  Computational Linguistics for Cultural Heritage, Social Sciences, Humanities
  and Literature}, pages 1--11. Association for Computational Linguistics.

\bibitem[{Parker et~al.(2011)Parker, Graff, Kong, Chen, and
  Maeda}]{Gigaword:11}
Robert Parker, David Graff, Junbo Kong, Ke~Chen, and Kazuaki Maeda. 2011.
\newblock {English Gigaword Fifth Edition LDC2011T07}.
\newblock Technical report, Linguistic Data Consortium, Philadelphia.

\bibitem[{Pettersson and Eck(2018)}]{pettersson2018organized}
Ther{\'e}se Pettersson and Kristine Eck. 2018.
\newblock Organized violence, 1989--2017.
\newblock \emph{Journal of Peace Research}, 55(4):535--547.

\bibitem[{Rogers et~al.(2017)Rogers, Drozd, and Li}]{rogers:analogy:2017}
Anna Rogers, Aleksandr Drozd, and Bofang Li. 2017.
\newblock \href {https://doi.org/10.18653/v1/S17-1017} {The (too many) problems
  of analogical reasoning with word vectors}.
\newblock In \emph{Proceedings of the 6th Joint Conference on Lexical and
  Computational Semantics (*SEM 2017)}, pages 135--148. Association for
  Computational Linguistics.

\bibitem[{Rosenfeld and Erk(2018)}]{rosenfeld:2018}
Alex Rosenfeld and Katrin Erk. 2018.
\newblock \href {http://aclweb.org/anthology/N18-1044} {Deep neural models of
  semantic shift}.
\newblock In \emph{Proceedings of the 2018 Conference of the North American
  Chapter of the Association for Computational Linguistics: Human Language
  Technologies}, pages 474--484, New Orleans, Louisiana, USA.

\bibitem[{Straka and Strakov{\'a}(2017)}]{udpipe:2017}
Milan Straka and Jana Strakov{\'a}. 2017.
\newblock \href {https://doi.org/10.18653/v1/K17-3009} {Tokenizing, pos
  tagging, lemmatizing and parsing {UD} 2.0 with {UDP}ipe}.
\newblock In \emph{Proceedings of the CoNLL 2017 Shared Task: Multilingual
  Parsing from Raw Text to Universal Dependencies}, pages 88--99. Association
  for Computational Linguistics.

\bibitem[{Tang(2018)}]{tang_2018}
Xuri Tang. 2018.
\newblock \href {https://doi.org/10.1017/S1351324918000220} {A state-of-the-art
  of semantic change computation}.
\newblock \emph{Natural Language Engineering}, 24(5):649–676.

\end{thebibliography}
\bibliographystyle{acl_natbib}

\end{document}